%% file: main.tex
\documentclass{article}

\usepackage[T1]{fontenc}  

\PassOptionsToPackage{sort&compress}{natbib}

\PassOptionsToPackage{hyperfootnotes=false}{hyperref}

\usepackage[final]{corl_2026} 

\usepackage{amsmath, amssymb}
\usepackage{booktabs}

\usepackage{multirow}
\usepackage{graphicx}
\usepackage{colortbl} 
\definecolor{lightgreen}{rgb}{0.82,0.94,0.82}
\definecolor{lightred}{rgb}{0.98,0.85,0.85}
\newcommand{\cmark}{\cellcolor{lightgreen}\checkmark}
\newcommand{\xmark}{\cellcolor{lightred}\ensuremath{\times}}

\title{PhysCoRe: Physics-Corrected Residual World Models for Material-Aware Deformable Dynamics}

\author{
  Haocheng Yin\thanks{Equal contribution.} \quad
  Shuohan Tao\footnotemark[1] \quad
  Yongsheng Chen \quad
  Lu Gan \\
  Georgia Institute of Technology
}

\begin{document}
\maketitle


\begin{abstract}
  Predicting how deformable objects evolve under robotic manipulation is a longstanding challenge. Existing approaches typically rely on per-object optimization to fit material parameters, which can be slow and cannot generalize, while end-to-end learned alternatives extrapolate poorly and often violate basic physical structure. We present PhysCoRe, a physics-corrected residual world model that couples a differentiable Material Point Method (MPM) simulator with two feed-forward neural networks. A material refinement module, \emph{Material from Motion} (MfM), infers per-particle elasticity from visual observations, grounding the simulator in object-specific physics. A residual correction module, \emph{Residual from Dynamics} (RfD), learns the discrepancy and predicts corrections to the simulator's internal dynamics, absorbing systematic biases that the analytical model cannot capture. This design also supports online material identification on novel objects. MfM adapts from limited interactions, and its predictive uncertainty steers further exploration toward the regions where its estimate is least confident. Experiments on real deformable-object manipulation sequences show that PhysCoRe outperforms state-of-the-art baselines in prediction accuracy, and that its predicted confidence forms a reliable distribution across the object's geometry, providing a natural signal for future confidence-guided exploration. Project page: \url{https://lunarlab-gatech.github.io/PhysCoRe-website/}
\end{abstract}


\input{sections/introduction}

\input{sections/related_work}

\input{sections/preliminary}

\input{sections/method}

\input{sections/experiments}

\input{sections/conclusion}

\input{sections/limitations}


\clearpage
\acknowledgments{This work was supported in part by the U.S. National Science Foundation under Grant Nos.~2345543 and 2112533, and by the U.S. Department of Agriculture's National Institute of Food and Agriculture under Grant Nos.~2024-67021-43862 and 2024-67021-41534. We thank the FLAIR Lab for providing the robot hardware used in our experiments.}


\bibliography{references}  


\clearpage
\appendix
\input{sections/appendix}

\end{document}

%% file: sections/introduction.tex
\section{Introduction}
\label{sec:introduction}
 
To manipulate a deformable object with intent, a robot must anticipate how that object will move before committing to an action. Such anticipation relies on a dynamics model that maps a planned interaction to the resulting deformation, and its accuracy limits how good the plan can be. This requirement is particularly demanding for deformable objects. Their behavior is governed by underlying material properties, for example stiffness and plasticity. These properties differ from one object to the next and can vary even within a single body, and none of them is directly readable from an RGB-D observation. A useful dynamics model must therefore infer this latent material behavior from visual observations and remain accurate as the object evolves into configurations not encountered during training.

\begin{figure}[t]
    \centering
    \includegraphics[width=\linewidth]{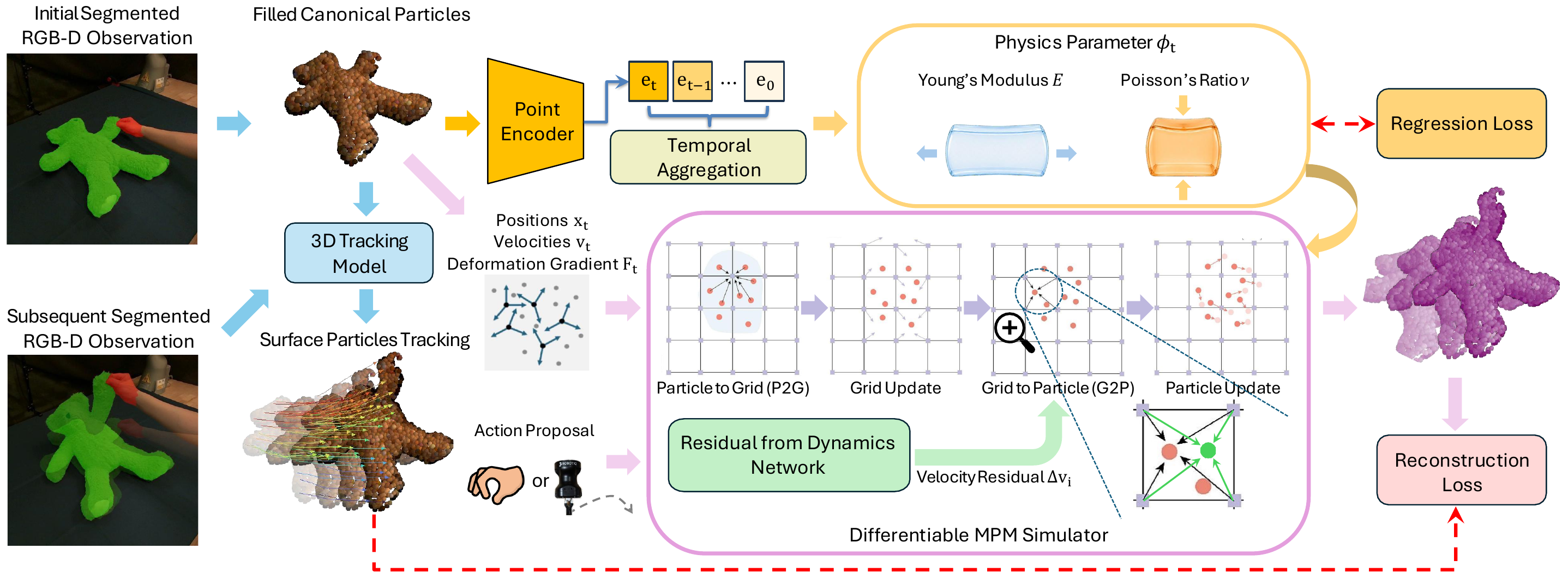}
    \caption{\textbf{Overview of PhysCoRe.} From segmented RGB-D observations, MfM infers the object's per-particle material. A differentiable MPM simulator then rolls out the action-conditioned dynamics, with RfD injecting a velocity residual $\Delta\mathbf{v}_i$ to correct the grid velocity. The two modules are supervised by a material regression loss and a reconstruction loss, respectively.}
    \label{fig:pipeline}
\end{figure}

Recent years have seen rapid progress in modeling the dynamics of deformable objects, through physical simulators, data-driven predictors, and combinations of the two~\citep{hu2018moving, jiang2025phystwin, sanchez2020learning, pfaff2021learning, zhang2025particle, chen2026empm}. Even so, combining accuracy, physical consistency, and broad generalization within a single model remains challenging. Physics-based methods produce physically plausible motion, but they rely on material parameters that are hard to measure and must be recalibrated for each new object. Data-driven methods avoid this calibration, yet they are data-hungry and not guaranteed to respect physical law. As a result, no existing approach identifies material quickly enough to transfer across objects while keeping its predictions grounded in physics.

In this work, we propose PhysCoRe, a dynamics model that retains a differentiable MPM simulator at its core and augments it with two learned components. The first replaces per-object calibration with inference: from a short window of observed motion, it estimates the object's material directly, in a single forward pass that transfers to objects never seen before. The second targets the residual error that any analytical simulator leaves behind, since coarse discretization and idealized material laws keep it from matching reality exactly. The material inference additionally outputs a per-particle confidence. We observe that this confidence varies coherently over the object's shape, marking where its estimate is dependable, and it can later serve to direct confidence-guided exploration.

We evaluate PhysCoRe on real manipulation sequences spanning elastic and elastoplastic objects. It predicts movement and deformation more accurately than state-of-the-art physics-based and learned baselines, and its predicted confidence concentrates on the parts of the object that deform most.

In summary, our contributions are as follows:

\begin{itemize}
    \item \textbf{Material from Motion (MfM):} a feed-forward module that infers per-particle material parameters and an associated confidence directly from visual observations, grounding a differentiable MPM simulator in object-specific physics and generalizing to unseen objects.
    \item \textbf{Residual from Dynamics (RfD):} a feed-forward module that corrects the physics simulator's internal dynamics within the MPM cycle, accounting for the systematic sim-to-real errors the simplified model introduces while preserving its physical structure.
    \item \textbf{Confidence-guided active exploration:} we demonstrate that the per-particle confidence from MfM aligns with the deformation each region undergoes, so it reveals where the material estimate holds and supplies the signal that future active exploration would rely on.
\end{itemize}

%% file: sections/related_work.tex
\section{Related Work}
\label{sec:related_work}

\subsection{Material Identification and Dynamics Modeling for Deformable Objects}
\label{subsec:material_dynamics}
Modeling deformable-object dynamics has been pursued along two largely separate lines: physics-based modeling and learning-based modeling. The first line represents the object with an analytical model, such as a spring-mass system~\citep{provot1995deformation, liu2013fast}, the Finite Element Method (FEM)~\citep{garcia2006optimized, sifakis2012fem}, Position-Based Dynamics (PBD)~\citep{muller2007position}, or the Material Point Method (MPM)~\citep{sulsky1995application, jiang2016material, hu2018moving}, and recovers its material parameters from observed motion through differentiable simulation or gradient-free optimization. Recent methods instantiate this paradigm across 3D Gaussian representations~\citep{jiang2025phystwin, zhong2024reconstruction, lin2025omniphysgs}, neural constitutive laws~\citep{ma2023learning}, and continuum materials~\citep{jatavallabhula2021gradsim, gao2025seeing, li2023pacnerf, chen2022virtual, yang2025differentiable}, while others instead infer material properties from learned priors~\citep{zhang2024physdreamer, huang2025dreamphysics, liu2024physics3d, lv2026physgm}. These methods are interpretable and physically plausible, but identification is typically slow, object-specific, and constrained by the simulator's discretization and constitutive assumptions. The second line, surveyed in~\citep{ai2025review}, learns dynamics directly from interaction data, from graph neural networks~\citep{battaglia2016interaction, li2019learning, pfaff2021learning, sanchez2020learning} to diverse architectures spanning plasticine, cloth, rope, and granular media~\citep{lin2021learning, ma2022learning, shi2023robocook, wang2023dynamic, shi2024robocraft, zhang2024dynamics, zhang2025particle, xue2023neural, tian2025uniclothdiff, bauer2024doughnet, yan2020self, orozco2026learning, whitney2024learning, whitney2024modeling}, with several adapting online from limited interaction~\citep{zhang2024adaptigraph, longhini2023edonet, wang2022offline}. Purely data-driven variants, however, require large datasets, generalize poorly out of distribution, and lack physical structure. Our work seeks to combine the strengths of both lines, retaining a physical simulator to advance the dynamics while delegating the unknown material properties to a feed-forward neural module that infers per-particle elasticity from visual observations, and thereby generalizes to previously unseen deformable objects from limited interaction.

\subsection{World Models for Object Dynamics}
\label{subsec:world_models}
World models predict how an object will evolve under a planned action and are distinguished by their state representation. Pixel-space video models predict future frames, while structured 3D approaches operate on particles or Gaussians~\citep{huang2026pointworld, huang2025particleformer, daniel2026latent, kipf2020contrastive, zhou2025dinowm, lu2025gwm}. Recent works pursue training from autonomous robot data~\citep{yin2026playworld}, scaling across embodiments~\citep{he2025scaling}, fusing with scene reconstruction~\citep{driess2022learning} or tactile sensing~\citep{ai2024robopack}, and guiding manipulation and planning at inference time~\citep{qi2026gpc, liu2023model}. These models predict future state by tracking 2D pixels or 3D points without an analytical physics simulator. That suffices for rigid objects but leaves deformable ones underdetermined, since the same observation and action sequence can reach very different configurations depending on material properties that observations do not reveal. This motivates embedding physical structure into the representation~\citep{jiang2025phystwin, abou2024embodied, chen2026empm, abou2025real, patel2026pgrd}, a direction our work follows by organizing the world model around a physics simulator grounded in inferred material properties, so that action-conditioned predictions remain physically consistent.

\subsection{Differentiable Physics Simulation for Deformable Objects}
\label{subsec:diff_sim}
Differentiable physics simulation has advanced rapidly for deformable objects, with MPM providing an effective discretization for continuum materials under large deformation, fracture, and contact~\citep{jiang2015affine, hu2018moving}. Differentiable solvers serve as instruments for system identification and gradient-based control~\citep{hu2019chainqueen, schenck2018spnets, su2023generalized, mittal2025uniphy} as well as benchmarks for soft-body manipulation~\citep{huang2021plasticinelab}. Dynamic scene reconstruction methods recover appearance and geometry of deformable objects from video~\citep{luiten2024dynamic, pumarola2021dnerf, wu2024gaussian, yang2024deformable, longhini2024cloth}, and physics-grounded variants augment these representations with differentiable simulators that also predict the underlying dynamics~\citep{xie2024physgaussian, huang2026gaussianfluent, chen2026empm}. However, an analytical simulator never matches reality exactly: coarse discretization, simplified constitutive laws, and unmodeled contact and friction leave a persistent sim-to-real gap. This observation motivates residual-physics methods, which train a learned module to predict a correction that compensates the simulator output. Such corrections have been applied to rigid robots~\citep{golemo2018sim}, planar pushing under uncertainty~\citep{ajay2018augmenting}, high degree of freedom soft robots~\citep{gao2024sim}, and more recently to deformable objects~\citep{patel2026pgrd}. Our work adopts this paradigm to close the residual gap of a differentiable MPM simulator, training a correction module on its internal dynamics to recover the behavior the analytical model fails to reproduce perfectly.

%% file: sections/preliminary.tex
\section{Preliminary}
\label{sec:preliminary}

We adopt the Moving Least Squares Material Point Method (MLS-MPM)~\citep{hu2018moving} with Affine Particle-in-Cell (APIC) transfers~\citep{jiang2015affine} as our differentiable simulator, discretizing the object as $N$ Lagrangian particles that exchange momentum with a fixed background grid~\citep{sulsky1995application, jiang2016material}. The dynamics are integrated in explicit time substeps of size $\Delta t$ indexed by $h$, with $H$ substeps per camera frame. Each substep advances the particle state through the following four steps. The full constitutive and plasticity models, boundary conditions, and simulator parameters are described in Appendix~\ref{app:simulator}.

\paragraph{Particle to grid (P2G).}
Each particle $p$ carries mass $m_p$, velocity $\mathbf{v}_p$, and an APIC affine matrix $\mathbf{C}_p \in \mathbb{R}^{3 \times 3}$, and couples to its surrounding grid nodes through quadratic B-spline weights $w_{p,i}$ (with gradient $\nabla w_{p,i}$) and the offset $\mathbf{d}_{p,i} = \mathbf{x}_i - \mathbf{x}_p$ from particle position $\mathbf{x}_p$ to grid-node position $\mathbf{x}_i$. The substep first scatters mass and momentum, together with the internal-force impulse from the Cauchy stress $\boldsymbol{\sigma}_p$, to grid node $i$,
\begin{equation}
    m_i = \sum_{p} w_{p,i}\,m_p, \qquad
    (m\mathbf{v})_i = \sum_{p} \Bigl[\, w_{p,i}\,m_p\,(\mathbf{v}_p^{h} + \mathbf{C}_p^{h}\,\mathbf{d}_{p,i}) - \Delta t\,V_p\,\boldsymbol{\sigma}_p\,\nabla w_{p,i} \,\Bigr],
    \label{eq:p2g}
\end{equation}
where $m_i$ and $(m\mathbf{v})_i$ are the resulting grid mass and momentum, and $V_p$ is the particle volume. The stress $\boldsymbol{\sigma}_p$ follows the \emph{Fixed Corotated Elasticity} constitutive model, evaluated from the per-particle material $\boldsymbol{\phi}_p = (\log E_p, \nu_p)$. Our MfM module infers this material from visual observations and is introduced in Section~\ref{subsec:mfm}.

\paragraph{Grid update.}
Each node recovers its velocity $\mathbf{v}_i$ from the scattered momentum, integrates gravity $\mathbf{g}$, and is damped by $\alpha \in (0, 1]$,
\begin{equation}
    \mathbf{v}_i = \alpha\,\Bigl(\frac{(m\mathbf{v})_i}{m_i} + \Delta t\,\mathbf{g}\Bigr).
    \label{eq:grid_update}
\end{equation}
Boundary conditions are then applied to $\mathbf{v}_i$, with a gripper or finger tip prescribing the velocity of contacting cells and the ground enforcing frictional contact. Our proposed RfD module is designed to add a learned correction to node velocity before the G2P gather, as detailed in Section~\ref{subsec:rfd}.


\paragraph{Grid to particle (G2P).}
The updated grid velocity is gathered back to the new particle velocity and APIC matrix over the same stencil,
\begin{equation}
    \mathbf{v}_p^{h+1} = \sum_{i} w_{p,i}\,\mathbf{v}_i, \qquad
    \mathbf{C}_p^{h+1} = \frac{4}{\Delta x^2}\,\sum_{i} w_{p,i}\,\mathbf{v}_i \otimes \mathbf{d}_{p,i},
    \label{eq:g2p}
\end{equation}
where $\Delta x$ is the grid cell size, $4/\Delta x^2$ is the APIC normalization for the quadratic kernel, and $\otimes$ denotes the outer product.

\paragraph{Particle update.}
Finally, the particle position advects and its deformation gradient $\mathbf{F}_p \in \mathbb{R}^{3 \times 3}$ is updated by the local velocity gradient, then the plasticity return map $\mathcal{P}(\cdot)$ is applied,
\begin{equation}
    \mathbf{x}_p^{h+1} = \mathbf{x}_p^{h} + \Delta t\,\mathbf{v}_p^{h+1}, \qquad
    \mathbf{F}_p^{h+1} = \mathcal{P}\!\Bigl(\bigl(\mathbf{I} + \Delta t \sum_{i} \mathbf{v}_i \otimes \nabla w_{p,i}\bigr)\,\mathbf{F}_p^{h} \Bigr).
    \label{eq:particle_update}
\end{equation}
$\mathcal{P}$ either leaves $\mathbf{F}_p$ unchanged (Fixed Corotated Elasticity) or projects it onto the \emph{von Mises Plasticity} yield surface; MfM selects the branch per episode. The discrete action sequence $\mathbf{a}_{0:T-1}$, defined per camera frame, is temporally upsampled to the substep rate by a Catmull--Rom spline~\citep{catmull1974class}.

%% file: sections/method.tex
\section{Method}
\label{sec:method}

\subsection{Problem Formulation}
\label{subsec:problem}

Given an initial multi-view RGB-D observation $\mathbf{O}_0$ of a deformable object and a proposed action sequence $\mathbf{a}_{0:T-1}$ from a robot gripper or human hand, we learn a physics-based dynamics model that predicts the future particle configurations $\mathbf{X}_{1:T}$. Here $\mathbf{X}_t = \{\mathbf{x}_p^{(t)}\}_{p=1}^{N}$ are the positions of the object's $N$ continuum particles at camera frame $t$. Since the governing material properties are not directly observable, the model infers them from the observations and parameterizes the simulator accordingly, so that action-conditioned predictions stay physically consistent.

\subsection{PhysCoRe Framework}
\label{subsec:framework}

We build on the differentiable MPM simulator of Section~\ref{sec:preliminary}, inheriting its particle-grid state, B-spline kernel, constitutive stress model, and plasticity return map, and couple it with two learned modules. \emph{Material from Motion} (MfM) infers per-particle material parameters and a per-particle confidence from visual observations, together with a constitutive-model probability that selects between Fixed Corotated Elasticity and von Mises Plasticity. \emph{Residual from Dynamics} (RfD) then reads features from the current grid state and MfM's outputs and emits a bounded grid-velocity residual $\Delta\mathbf{v}_i$, added to the analytical grid velocity before G2P to absorb discrepancies between the idealized simulator and the real dynamics.

\subsubsection{Material from Motion (MfM) Module}
\label{subsec:mfm}

MfM estimates an object's material from a short window of observed motion. The canonical positions and the per-frame displacements of the tracked object and controller points are encoded via Fourier features. The encoded features feed a graph U-Net~\citep{gao2019graph} that gives more weight to the tracked points and the particles near the controllers during message passing, since these particles are the most informative about the material. A temporal module then combines a depthwise one-dimensional convolution over the frames with gated recurrent unit (GRU) cells to produce a single per-particle latent. A shared MLP decodes this latent into the per-particle material $\boldsymbol{\phi}_p = (\log E_p, \nu_p)$, using a bounded sigmoid to keep Young's modulus and Poisson's ratio inside a fixed range that covers the training distribution and keeps the simulator numerically stable. It also predicts a per-particle confidence $\mathbf{c}_p$, with one value for Young's modulus and one for Poisson's ratio. This confidence reweights the material loss during training (Section~\ref{subsec:training}) and indicates the regions where MfM's estimate is most reliable at inference. Separately, the per-particle latent is pooled across all particles and fed to a smaller two-layer MLP that outputs a per-episode probability $\pi$ of elastic versus plastic behavior. We give the full architecture in Appendix~\ref{app:mfm}.

\subsubsection{Residual from Dynamics (RfD) Module}
\label{subsec:rfd}

RfD closes the residual sim-to-real gap left by the analytical simulator. It acts inside the MLS-MPM cycle, between the grid update of Eq.~\eqref{eq:grid_update} and the G2P gather of Eq.~\eqref{eq:g2p}. For each active cell it builds a compact feature $\mathbf{f}_i$ that gathers the local grid state together with the MfM material estimate and the nearby contact geometry. A FiLM-conditioned~\citep{perez2018film} sparse 3D U-Net~\citep{cicek2016unet3d, graham2018sparse} then maps this feature to a grid-velocity residual,
\begin{equation}
    \Delta\mathbf{v}_i = \delta_{\max} \cdot \tanh\!\Bigl(\mathcal{R}_\psi(\mathbf{f}_i;\, \mathbf{u})\Bigr),
    \label{eq:residual_update}
\end{equation}
where $\mathbf{u}$ is a global context vector that summarizes the current simulation regime. The scaled $\tanh$ limits every component of $\Delta\mathbf{v}_i$ to a maximum magnitude $\delta_{\max}$, so the learned correction stays small enough not to destabilize the rollout. We zero-initialize the network's output layer, so the residual starts at exactly zero and RfD begins as an identity correction that leaves the analytical simulator unchanged. Because the correction is applied on the grid and carried to the particles by the same APIC transfer, it preserves the momentum and angular-momentum structure of MLS-MPM to first order. The corrected velocity $\mathbf{v}'_i = \mathbf{v}_i + \Delta\mathbf{v}_i$ then replaces the analytical velocity $\mathbf{v}_i$ in the G2P gather of Eq.~\eqref{eq:g2p} and the particle update of Eq.~\eqref{eq:particle_update}. RfD is applied periodically rather than at every substep, which bounds the backpropagation graph while still exposing it to the error accumulated between successive corrections. We give the full feature and network details in Appendix~\ref{app:rfd}.

\subsubsection{Material Augmentation}
\label{subsec:material_augmentation}
Supervised material data is scarce, since measuring the spatially varying stiffness and Poisson's ratio of everyday objects is impractical at scale. We therefore generate this supervision synthetically, inspired by LRM-Zero~\citep{xie2024lrmzero}. Starting from 14 real PhysTwin episodes~\citep{jiang2025phystwin}, we build $1{,}120$ augmented episodes. Each one inherits the initial geometry and the full controller trajectory of its source capture, and follows a different deformation rollout under its augmented material.

For each episode we draw a spatially correlated material field over the canonical particles $\mathbf{X}_0$ from multi-octave Perlin noise and affinely map it to per-particle parameters through episode-level mean and scale statistics,
\begin{equation}
    \log E_p = \mu_{\log E} + \sigma_{\log E}\,z_p^{(\log E)},
    \qquad
    \nu_p = \mu_{\nu} + \sigma_{\nu}\,z_p^{(\nu)}.
    \label{eq:material_field}
\end{equation}
Here $z_p^{(\log E)}$ and $z_p^{(\nu)}$ are independent standardized noise fields evaluated at particle $p$. Each episode also carries a plasticity label $\pi^{\star}$, assigned when the episode is specified, which selects the constitutive model for that episode. Elastic episodes use Fixed Corotated Elasticity alone, while plastic episodes add a von Mises return map. We then roll out the MPM simulator from $\mathbf{X}_0$ under the capture's real controller action trajectory, so the augmented motion is driven by the same boundary forcing as the original. Finally, we re-render the resulting rollout from the saved camera views to produce the per-frame point clouds $\hat{\mathbf{Y}}_t$ and particle correspondences that supervise training in Section~\ref{subsec:training}. Sampling ranges and rendering details are provided in Appendix~\ref{app:augmentation}.

\subsubsection{Two-Stage Training}
\label{subsec:training}

We train the modules sequentially: MfM is pre-trained in simulation against ground-truth materials on the augmented dataset of Section~\ref{subsec:material_augmentation}, then RfD is trained on real episodes with MfM frozen, supervising the rolled-out state against observations.

\paragraph{MfM pre-training.}
With ground-truth materials $\boldsymbol{\phi}^{\star}_p = (\log E^{\star}_p, \nu^{\star}_p)$ and label $\pi^{\star}$, the refinement loss is
\begin{equation}
    \mathcal{L}_{\mathrm{refine}} = \mathbb{E}\!\left[\frac{1}{N}\sum_{p=1}^{N}\sum_{k \in \{\log E,\,\nu\}}\!\Bigl(\,c_{p,k}\,\mathrm{SL1}\bigl(\bar{\phi}_{p,k},\, \bar{\phi}^{\star}_{p,k}\bigr) - \lambda \log c_{p,k} \Bigr)\right] + \beta\,\mathrm{BCE}\bigl(\pi, \pi^{\star}\bigr),
    \label{eq:loss_refiner}
\end{equation}
where $\bar{\phi}_{p,k}$ and $\bar{\phi}^{\star}_{p,k}$ are the predicted and ground-truth materials normalized to $[0,1]$, and $c_{p,k}$ is MfM's confidence in axis $k$ at particle $p$, which weights the corresponding Smooth-L1 error $\mathrm{SL1}$. The log-confidence regularizer keeps $c_{p,k}$ from inflating to trivially shrink the error, and the cross-entropy is class-weighted for the elastic-heavy imbalance, as described in Appendix~\ref{app:training}.

\paragraph{RfD training.}
With MfM frozen, RfD is trained end-to-end across MPM substeps. MfM first observes the opening half of the episode to produce a material estimate, which is then held fixed. Starting from the first observed frame, the MPM simulator is rolled out over that same half with RfD applying its correction every $H_{\mathrm{r}}$ substeps. At each frame the loss compares the predicted configuration $\mathbf{X}_t$ to the observations, combining a one-sided Chamfer distance from the point cloud $\hat{\mathbf{Y}}_t$ with a mean $L_2$ error on the visible tracked particles $\mathcal{T}_t$,
\begin{equation}
    \mathcal{L}_{\mathrm{corr}}^{(t)} = w_{\mathrm{C}}\,\mathrm{Chamfer}\bigl(\hat{\mathbf{Y}}_t,\, \mathbf{X}_t\bigr) + w_{\mathrm{L2}}\,\frac{1}{|\mathcal{T}_t|}\!\sum_{p \in \mathcal{T}_t}\!\bigl\|\mathbf{x}_p^{(t)} - \hat{\mathbf{x}}_p^{(t)}\bigr\|_2.
    \label{eq:loss_corr}
\end{equation}
Here $\hat{\mathbf{x}}_p^{(t)}$ is the observed position of tracked particle $p$ at frame $t$, and we set $w_{\mathrm{C}} = w_{\mathrm{L2}} = 1$. We sum this per-frame loss over each segment of the rollout and backpropagate through the enclosed substeps. Further training details are given in Appendix~\ref{app:training}.

%% file: sections/experiments.tex
\section{Experiments}
\label{sec:experiments}

\subsection{Experimental Setup}
Our real-world dataset consists of 12 episodes of deformable object manipulation by human hand. It covers elastic objects (a rope, a towel, and a plush toy bear) and elastoplastic objects (Play-Doh plasticine), manipulated through actions including lifting, pushing, stretching, and squeezing. Each episode is captured in RGB-D by three Intel RealSense D455 cameras. From every capture, CoTracker3~\citep{karaev2025cotracker3} provides tracked surface points and Grounded SAM2~\citep{ren2024grounded} provides object masks. All training and timing runs use a single RTX 5090 GPU. 

For the online confidence-guided exploration of Section~\ref{subsec:exploration} and the planning experiments of Section~\ref{subsec:planning}, we use a KUKA arm with a Robotiq Hand-E gripper.

We compare against PhysTwin~\citep{jiang2025phystwin} and PGND~\citep{zhang2025particle}. Table~\ref{tab:capability} also lists EMPM~\citep{chen2026empm}, whose code is unavailable for numerical comparison, and shows that only PhysCoRe spans all four capabilities.

\begin{table}[tb]
    \centering
    \small
    \caption{\textbf{Capability comparison.} PhysCoRe combines physics simulation, online material refinement, feed-forward inference, and generalization to unseen objects.}
    \label{tab:capability}
    \resizebox{0.7\textwidth}{!}{%
        \input{tables/tab_capability}
    }
\end{table}

\vspace{-5pt}
\subsection{Future Prediction}
Each method identifies its material from the first half of an episode (the identification window) and predicts the held-out remainder with it fixed, with PhysCoRe doing so feed-forward and the baselines by inference or optimization. We measure geometric accuracy with Chamfer distance (CD) and a tracking loss on the visible points, and visual quality with IoU, PSNR, SSIM, and LPIPS.

\begin{figure}[tb!]
    \centering
    \includegraphics[width=\linewidth]{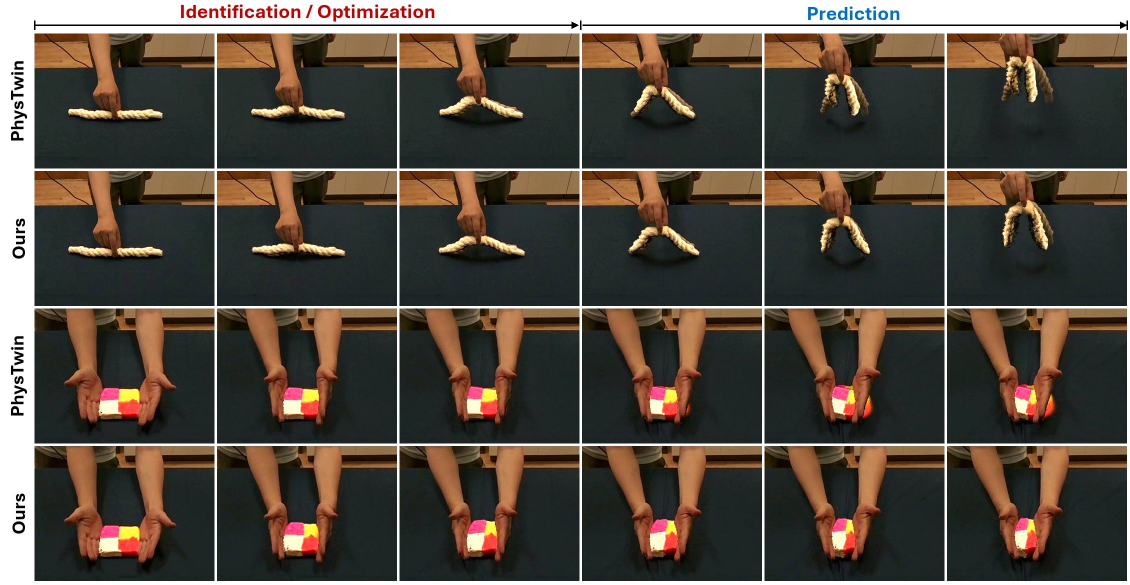}
    \caption{\textbf{Qualitative rollout comparison.} Predicted object positions are rendered with 3DGS. PhysCoRe predicts future dynamics closer to the real observations than PhysTwin.}
    \label{fig:qualitative}
    \vspace{-4pt}
\end{figure}

As Table~\ref{tab:future_pred} shows, PhysCoRe attains the best overall accuracy. Against PhysTwin it reduces CD by 43.7\% on elastic objects and 30.5\% on elastoplastic objects, and tracking loss by 15.2\% and 8.3\%, showing that grounding MPM with MfM-inferred materials yields more accurate geometry and point-level motion. The gains are most consistent on elastoplastic objects, where the simulator faces the widest sim-to-real gap. On elastic objects, PhysCoRe leads on geometry and stays competitive on visual quality, indicating that a single feed-forward inference can match per-object optimization even where the baseline is already strong. Figure~\ref{fig:qualitative} shows the same trend qualitatively, with the predicted object positions rendered by 3D Gaussian Splatting (3DGS)~\citep{kerbl2023gaussian}. PhysCoRe's rollouts stay closer to the observations as the horizon lengthens. Feed-forward inference is also far cheaper, identifying material in 11.4\,s against 930.0\,s for PhysTwin and 8280.0\,s for PGND (Table~\ref{tab:runtime}), which makes online adaptation to new objects practical. At test time PhysCoRe rolls out an episode in 8.5\,s, about 110\,ms per predicted frame. PGND is faster in this column because a single network forward pass replaces the physics simulation entirely, at a substantial cost in accuracy.

\begin{table}[tb!]
    \centering
    \small
    \caption{\textbf{Quantitative results on future prediction.} PhysCoRe achieves the best overall performance across both object types.}
    \label{tab:future_pred}
    \resizebox{\textwidth}{!}{%
        \input{tables/tab_future_pred}
    }
\end{table}
\subsection{Ablation: Effect of Residual Correction}
Table~\ref{tab:rfd_ablation} isolates RfD by comparing the MfM-only and full models. MfM alone is already a strong physics-grounded predictor, and adding RfD further reduces CD by 13.1\% on elastic objects and 17.8\% on elastoplastic objects. The larger elastoplastic gain confirms the intended division of labor: MfM identifies the object-specific material, while RfD absorbs the residual errors that simplified analytical MPM leaves from contact, friction, and irreversible deformation, without replacing it.

\begin{table}[tb!]
    \centering
    \begin{minipage}[t]{0.60\linewidth}
        \vspace{0pt}
        \centering
        \setlength{\belowcaptionskip}{3pt}
        \caption{\textbf{Ablation study on the effect of RfD.} RfD closes the residual sim-to-real gap left by the analytical simulator.}
        \label{tab:rfd_ablation}
        {\scriptsize
            \input{tables/tab_rfd_ablation}
        }
        \vspace{4pt}
        \caption{\textbf{Quantitative comparison on planning performance.} PhysCoRe achieves the best result.}
        \label{tab:planning}
        {\scriptsize
            \input{tables/tab_planning}
        }
    \end{minipage}\hfill
    \begin{minipage}[t]{0.37\linewidth}
        \vspace{0pt}
        \centering
        \caption{\textbf{Runtime comparison with baselines.} Identification/optimization time is the cost of estimating an object's material, and testing time is the rollout time. Both are measured per episode, averaged over the 12 evaluation episodes, and reported in seconds.}
        \label{tab:runtime}
        {\scriptsize
            \input{tables/tab_runtime}
        }
    \end{minipage}
\end{table}

\subsection{Confidence for Material Identification}
\label{subsec:confidence}
\begin{figure}[tb!]
    \centering
    \includegraphics[width=\linewidth]{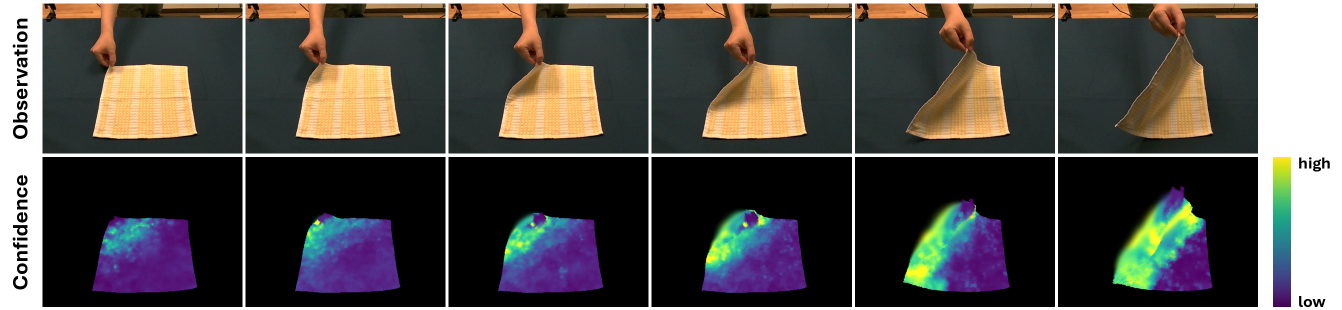}
    \caption{\textbf{MfM confidence during material identification.} The per-particle confidence is rendered onto the object with 3DGS. As more motion is observed, MfM becomes more confident in its material estimates. Brighter colors indicate higher confidence.}
    \label{fig:conf_validation}
\end{figure}

In addition to predicting the material itself, MfM outputs a per-particle confidence. Figure~\ref{fig:conf_validation} renders this confidence onto the object with 3DGS. As more of an object's motion is observed, MfM grows more confident, and its high-confidence region concentrates on the parts that have actually deformed. The confidence therefore tracks how well each region's material has been identified, and it emerges without any supervision beyond the log-regularized material loss of Eq.~\eqref{eq:loss_refiner}.

\subsection{Confidence-Guided Exploration}
\label{subsec:exploration}
\begin{figure}[tb!]
    \centering
    \includegraphics[width=\linewidth]{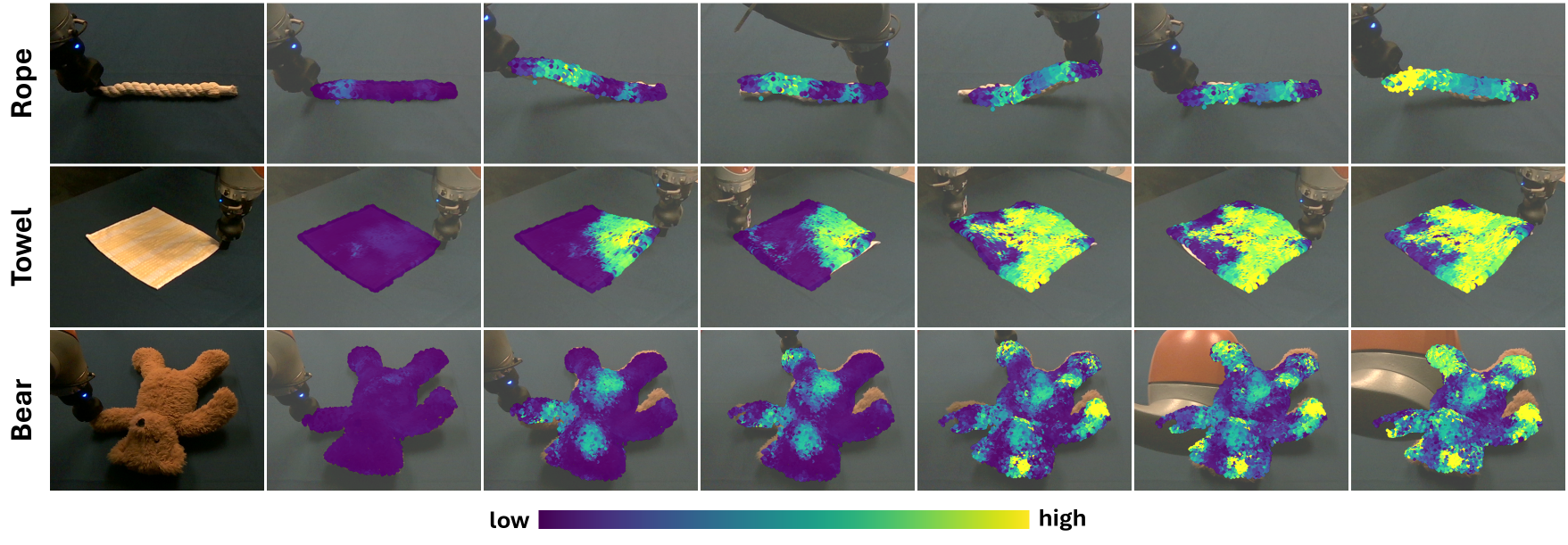}
    \caption{\textbf{Confidence-guided exploration on a real robot.} As a KUKA arm probes each object, MfM confidence starts low everywhere and rises only in regions the arm has actively deformed. It stays low elsewhere, so the map indicates where further interaction is most informative. Brighter colors indicate higher confidence.}
    \label{fig:conf_guidance}
\end{figure}

The same confidence supplies a signal for active exploration. In a real-world setup, we use a KUKA arm to probe a rope, a towel, and a plush toy, and render the resulting confidence maps, shown in Figure~\ref{fig:conf_guidance}. Confidence starts low across each object and rises in a region only after the arm has actively deformed it, and stays low elsewhere. The map therefore points at where interaction is still needed, which is the signal an active exploration policy would follow.

\subsection{Goal-Conditioned Planning}
\label{subsec:planning}
\begin{figure}[tb!]
    \centering
    \includegraphics[width=\linewidth]{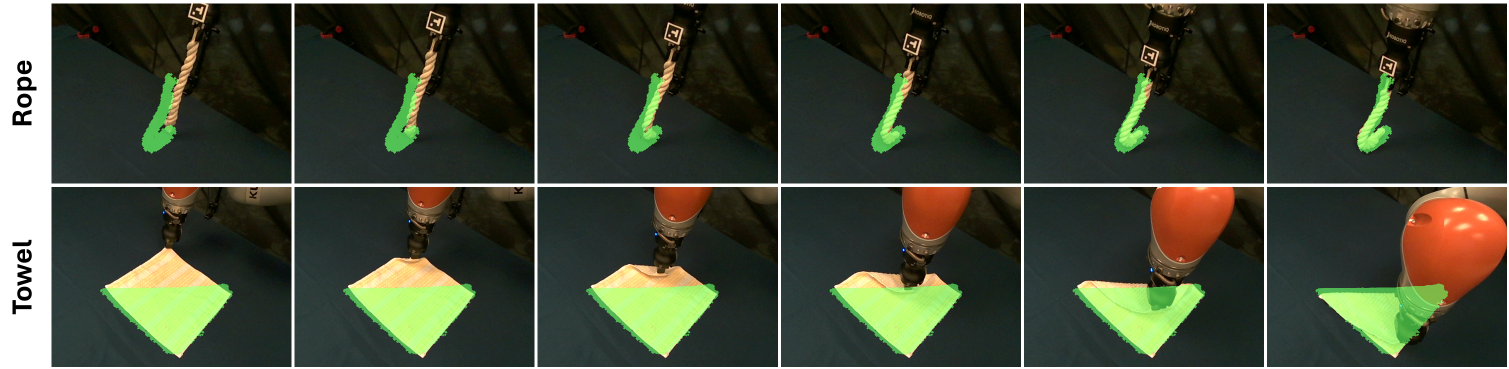}
    \caption{\textbf{Planning with PhysCoRe.} Green masks show the goal positions. Robot execution takes 9.0\,s on the rope and 23.5\,s on the towel.}
    \label{fig:planning}
\end{figure}

We next test whether more accurate prediction yields better plans. On a KUKA arm, we set up two manipulation tasks (Figure~\ref{fig:planning}): lowering a suspended rope to touch the table and shape half of an ``S'', and folding a towel along its diagonal. From a single RGB-D observation and a goal authored in the simulator, a gradient-based planner moves the gripper at each control step along the direction that reduces the predicted distance to the goal configuration the most. To isolate the dynamics model, we keep the planner, the goal, and all settings fixed and substitute only the backend with PhysTwin and PGND. Table~\ref{tab:planning} reports the symmetric Chamfer distance to the goal point cloud. PhysCoRe attains the lowest error on both tasks, showing that its more accurate forward prediction yields plans that leave the object closer to the goal configuration.

%% file: tables/tab_capability.tex
\begin{tabular}{lcccc}
\toprule
\textbf{Method} & \textbf{Simulation} & \textbf{Online Refinement} & \textbf{Feed-Forward} & \textbf{Generalization} \\
\midrule
PGND     & \xmark & \xmark & \cmark & \xmark \\
PhysTwin & \cmark & \xmark & \xmark & \xmark \\
EMPM     & \cmark & \cmark & \xmark & \xmark \\
PhysCoRe & \cmark & \cmark & \cmark & \cmark \\
\bottomrule
\end{tabular}

%% file: tables/tab_future_pred.tex
\begin{tabular}{lcccccccccccc}
\toprule
\multirow{2}{*}{\textbf{Method}} 
& \multicolumn{6}{c}{\textbf{Elastic Objects}} 
& \multicolumn{6}{c}{\textbf{Elastoplastic Objects}} \\
\cmidrule(lr){2-7} \cmidrule(lr){8-13}
& CD$\downarrow$ & Track$\downarrow$ & IoU\%$\uparrow$ & PSNR$\uparrow$ & SSIM$\uparrow$ & LPIPS$\downarrow$
& CD$\downarrow$ & Track$\downarrow$ & IoU\%$\uparrow$ & PSNR$\uparrow$ & SSIM$\uparrow$ & LPIPS$\downarrow$ \\
\midrule
PGND     
& 0.03420 & 0.13175 & 37.8 & 10.74 & 0.103 & 0.416
& 0.04311 & 0.13263 & 23.9 & 9.87 & 0.081 & 0.440 \\

PhysTwin 
& 0.01617 & 0.03016 & \textbf{70.5} & 14.15 & \textbf{0.309} & 0.199
& 0.00778 & 0.01821 & 63.1 & 14.16 & 0.285 & 0.154 \\

PhysCoRe     
& \textbf{0.00910} & \textbf{0.02557} & 68.2 & \textbf{14.47} & 0.302 & \textbf{0.194}
& \textbf{0.00541} & \textbf{0.01670} & \textbf{66.1} & \textbf{15.02} & \textbf{0.286} & \textbf{0.130} \\
\bottomrule
\end{tabular}

%% file: tables/tab_rfd_ablation.tex
\begin{tabular}{lcccc}
\toprule
\multirow{2}{*}{\textbf{Method}} & \multicolumn{2}{c}{\textbf{Elastic Objects}} & \multicolumn{2}{c}{\textbf{Elastoplastic Objects}} \\
\cmidrule(lr){2-3} \cmidrule(lr){4-5}
& CD$\downarrow$ & Track$\downarrow$ & CD$\downarrow$ & Track$\downarrow$ \\
\midrule
PhysCoRe (MfM)   & 0.01047 & 0.02723 & 0.00658 & 0.01981 \\
PhysCoRe (MfM + RfD) & 0.00910 & 0.02557 & 0.00541 & 0.01670 \\
\bottomrule
\end{tabular}

%% file: tables/tab_planning.tex
\begin{tabular}{lccc}
\toprule
\textbf{Chamfer (mm)} & PGND & PhysTwin & PhysCoRe (Ours) \\
\midrule
\textbf{Rope S-shaping} & 48.8 & 17.3 & 6.2 \\
\textbf{Towel folding}  & 30.6 & \phantom{0}7.1 & 5.3 \\
\bottomrule
\end{tabular}

%% file: tables/tab_runtime.tex
\begin{tabular}{lcc}
\toprule
\textbf{Time (s)} & \textbf{Id./Opt.} & \textbf{Testing} \\
\midrule
PhysTwin   & 930.0 & 7.6 \\
PGND & 8280.0 & 1.9 \\
PhysCoRe (Ours) & 11.4 & 8.5 \\
\bottomrule
\end{tabular}

%% file: sections/conclusion.tex
\section{Conclusion}
\label{sec:conclusion}


We presented PhysCoRe, a dynamics model that couples a differentiable MPM simulator with two feed-forward modules: MfM, which infers per-particle material from visual observations, and RfD, which corrects the simulator's internal dynamics. This design predicts deformable-object behavior accurately while retaining physical structure, and it transfers to unseen objects without per-object optimization. On real-world manipulation sequences, PhysCoRe outperforms state-of-the-art baselines in prediction accuracy, and its predicted confidence reliably reflects where the material estimate can be trusted, showing potential for confidence-guided exploration and active learning.

%% file: sections/limitations.tex
\section{Limitations}
\label{sec:limitations}



PhysCoRe focuses on elastic and elastoplastic objects. While MPM provides a flexible physics backbone for deformable dynamics, our implementation relies on a predefined set of material behaviors, so it may not extend to dynamics that involve tearing, cutting, adhesion, fluid-like motion, or highly complex contacts. Although our real-world experiments demonstrate fast adaptation to new objects and manipulation interactions, the evaluation is limited to in-category objects and does not test zero-shot category-level transfer. RfD is also trained on a limited set of deformable objects, and its scalability to broader dynamics distributions remains unexplored. 


%% file: sections/appendix.tex
\section*{Appendix}

\section{Simulator Details}
\label{app:simulator}

\paragraph{Constitutive stress.}
The Cauchy stress $\boldsymbol{\sigma}_p$ in Eq.~\eqref{eq:p2g} follows Fixed Corotated Elasticity. From the per-particle material $(\log E_p, \nu_p)$ we recover Young's modulus $E_p = \exp(\log E_p)$ and the Lam\'e parameters $\mu_p = E_p / (2(1+\nu_p))$ and $\lambda_p = E_p \nu_p / ((1+\nu_p)(1-2\nu_p))$. Writing the SVD $\mathbf{F}_p = \mathbf{U}_p \boldsymbol{\Sigma}_p \mathbf{V}_p^\top$ with $\mathbf{R}_p = \mathbf{U}_p \mathbf{V}_p^\top$ and $J_p = \det \mathbf{F}_p$,
\begin{equation}
    \boldsymbol{\sigma}_p = 2\mu_p\,(\mathbf{F}_p - \mathbf{R}_p)\,\mathbf{F}_p^\top + \lambda_p\,J_p (J_p - 1)\,\mathbf{I}.
    \label{eq:app_corotated}
\end{equation}
Particles tagged as kinematic (controller or rigid contact) have their stress zeroed so they move purely under the imposed velocity.

\paragraph{Plasticity return map.}
The map $\mathcal{P}(\cdot)$ in Eq.~\eqref{eq:particle_update} is either the identity (elastic regime) or a von Mises Plasticity projection. The plastic branch takes the SVD of the trial deformation gradient, forms the deviatoric Hencky strain $\boldsymbol{\epsilon}_{\mathrm{dev}}$ from $\log \boldsymbol{\Sigma}_p$, and, when $\|\boldsymbol{\epsilon}_{\mathrm{dev}}\|$ exceeds the yield threshold $\tau_y / (2\mu_p)$, scales the singular values back onto the yield surface.

\paragraph{Boundary conditions.}
Two velocity projections act on $\mathbf{v}_i$ after the grid update of Eq.~\eqref{eq:grid_update}. Kinematic handles and controller points splat a target velocity to nearby grid nodes and blend it in, injecting gripper or finger-tip motion. The ground uses a signed-distance Coulomb collider: for a node below the plane, the normal velocity is reflected with restitution $e$ and the tangential velocity is capped by Coulomb friction with coefficient $\mu$.

\paragraph{Default parameters.}
Table~\ref{tab:app_sim_params} lists the simulator settings used throughout.

\begin{table}[!ht]
\centering
\caption{\textbf{MLS-MPM simulator parameters.}}
\label{tab:app_sim_params}
\begin{tabular}{lll}
\toprule
Symbol & Meaning & Default \\
\midrule
$G$ & grid resolution ($\Delta x = 1/G$) & $32$ \\
$\Delta t$ & substep size & $6.66\times 10^{-4}$\,s \\
$H$ & substeps per camera frame & $50$ \\
$\mathbf{g}$ & gravity & $(0, 0, -9.8)$\,m/s$^2$ \\
$\alpha$ & velocity damping & $0.999$ \\
$\rho$ & density & $100$ \\
& kernel & quadratic B-spline ($3^3$ stencil) \\
& transfer & APIC (MLS-MPM) \\
& ground contact & Coulomb, $\mu = 0.5$, $e = 0$, height $0.02$ \\
\bottomrule
\end{tabular}
\end{table}

\section{MfM Architecture and Hyperparameters}
\label{app:mfm}

\paragraph{Input encoding.}
The observable blocks of Section~\ref{subsec:mfm} are encoded as in Table~\ref{tab:app_mfm_input}. Canonical positions and tracked displacements are Fourier-encoded, whereas the controller block is passed through unencoded. Rather than reserving one slot per controller, we pool all controller points into a single distance-weighted summary using an exponential kernel of falloff radius $0.04$\,m, so this block has a fixed width regardless of how many controllers act on the object. Concatenating the per-frame blocks gives $d_{\mathrm{in}} = 128$, which is stacked over the $K = 10$-frame window.

\begin{table}[tb]
\centering
\caption{\textbf{MfM per-frame input blocks.} Bands $B$ produce $6B$ Fourier features per 3-vector.}
\label{tab:app_mfm_input}
\begin{tabular}{ll}
\toprule
Block & Dim \\
\midrule
Canonical positions $\mathbf{x}_p^{(0)}$, mean-centered, Fourier with $B_c = 4$ & $24$ \\
Tracked displacement $\Delta\mathbf{x}_p^{(t)}$, Fourier with $B_t = 16$ & $96$ \\
Tracked visibility mask & $1$ \\
Controller vector (distance-weighted mean) & $3$ \\
Controller displacement (distance-weighted mean) & $3$ \\
Controller weight & $1$ \\
\bottomrule
\end{tabular}
\end{table}

\paragraph{Backbone.}
The spatial backbone is a graph U-Net with hidden width $d_h = 128$ and $L = 4$ message-passing layers. Its nodes are the object's particles, and at every level each node is joined to its $k = 8$ nearest neighbors in the canonical configuration, so the edges encode rest-state adjacency and stay fixed for the entire rollout. Each layer updates the per-particle feature from an evidence-weighted mean of its neighbor messages,
\begin{align}
    \mathbf{h}_p^{(\ell+1)} &= \eta_{\mathrm{u}}\!\Bigl(\mathbf{h}_p^{(\ell)},\; \frac{\sum_{q \in \mathcal{N}(p)} e_q\,\eta_{\mathrm{m}}\bigl(\mathbf{h}_q^{(\ell)}\bigr)}{\sum_{q \in \mathcal{N}(p)} e_q}\Bigr), \label{eq:mfm_update} \\
    e_q &= 1 + a_t\,s_q^{\mathrm{trk}} + a_c \max_{c}\, s_c\,\exp\!\bigl(-\|\mathbf{x}_q - \mathbf{z}_c\|/r_c\bigr), \label{eq:mfm_evidence}
\end{align}
where $\eta_{\mathrm{m}}$ is a learned message MLP and $\eta_{\mathrm{u}}$ a learned update MLP, $e_q$ is the neighbor evidence weight, $\mathcal{N}(p)$ is the neighbor set, $s_q^{\mathrm{trk}}$ and $s_c$ are visibility masks, and $\mathbf{z}_c$ is the $c$-th controller position. The evidence score uses $a_t = 8.0$, $a_c = 4.0$, and falloff radius $r_c = 0.04$\,m, and is pooled to coarse levels by $\max$. Temporal fusion applies a depthwise 1D convolution of kernel size $9$ across the frame axis, with a GRU cell on the upper half of the backbone ($\ell \ge L/2$). The recurrent state is threaded across windows within an episode and detached every four windows.

\paragraph{Output heads.}
The per-particle latent $\mathbf{h}_p$ is decoded by a shared four-layer MLP into raw material and confidence outputs $[\mathbf{r}_p^{\phi}, \mathbf{r}_p^{c}]$. A bounded sigmoid maps the material output into a fixed range,
\begin{equation}
    \boldsymbol{\phi}_p = \boldsymbol{\phi}_{\min} + (\boldsymbol{\phi}_{\max} - \boldsymbol{\phi}_{\min}) \odot \sigma\!\bigl(\mathbf{r}_p^{\phi}\bigr),
    \label{eq:mfm_output}
\end{equation}
where $\odot$ is the elementwise product and $\boldsymbol{\phi}_{\min}, \boldsymbol{\phi}_{\max}$ bound Young's modulus and Poisson's ratio at $\log E \in [5, 11]$ and $\nu \in [0.05, 0.45]$. The confidence output gives a per-axis $\mathbf{c}_p = 1 + \exp(\mathbf{r}_p^{c}) \in (1,\infty)^2$. A separate two-layer MLP $\eta_\pi$ with a sigmoid maps the max-pooled latent to the per-episode probability $\pi = \sigma(\eta_\pi(\max_p \mathbf{h}_p))$. The confidence and plasticity outputs are zero-initialized, giving $\mathbf{c}_p = 2$ and $\pi = \tfrac{1}{2}$ at the start of training, and $\pi$ is thresholded at $\tau = 0.5$.

\section{RfD Architecture and Hyperparameters}
\label{app:rfd}

\paragraph{Per-cell features.}
For each active grid node $i$ after P2G, RfD assembles the 19-dimensional feature
\begin{equation}
    \mathbf{f}_i = \bigl[\, \mathbf{v}_i,\; \log(1 + m_i),\; \mathrm{sym}(\overline{\boldsymbol{\sigma}}_i),\; \overline{\log E}_i,\; \overline{\nu}_i,\; \overline{\mathbf{c}}_i,\; \overline{\mathbf{v}}^{(p)}_i,\; \kappa_i,\; d_i^{\mathrm{ground}} \,\bigr] \in \mathbb{R}^{19}.
    \label{eq:rfd_features}
\end{equation}
The grid velocity $\mathbf{v}_i$ provides three channels and the log-compressed nodal mass $\log(1 + m_i)$ a single channel, while the symmetric Cauchy stress $\mathrm{sym}(\overline{\boldsymbol{\sigma}}_i)$ contributes its six independent components. The P2G-scattered MfM outputs occupy four channels, one each for the material parameters $\overline{\log E}_i$ and $\overline{\nu}_i$ and two for the per-particle confidence $\overline{\mathbf{c}}_i$. The scattered particle velocity $\overline{\mathbf{v}}^{(p)}_i$ accounts for three further channels. The remaining two are the contact indicator $\kappa_i$, which marks cells receiving mass from particles in contact with the gripper or finger tip, and the signed distance $d_i^{\mathrm{ground}}$ from the cell center to the ground plane. Each per-particle quantity $a_p \in \{\boldsymbol{\sigma}_p, \log E_p, \nu_p, \mathbf{c}_p, \mathbf{v}_p\}$ is carried to grid node $i$ by the same mass-weighted P2G transfer used for momentum,
\begin{equation}
    \bar{a}_i = \frac{1}{m_i}\sum_{p} w_{p,i}\,m_p\,a_p,
    \label{eq:rfd_scatter}
\end{equation}
so the scattered inputs occupy exactly the active cells of the simulator's momentum scatter, avoiding any mismatch between the simulator and RfD. The stress components are compressed elementwise by $\mathrm{sign}(s)\log(1 + |s|)$ before this scatter, bounding their dynamic range.

\paragraph{Backbone and conditioning.}
The backbone is a sparse 3D U-Net of submanifold sparse convolutions (which preserve the active set), with stride-2 sparse convolutions for downsampling and inverse sparse convolutions for upsampling. Every block applies its sparse convolution, modulates the result by a FiLM scale and shift, and passes it through a ReLU,
\begin{align}
    \bigl[\boldsymbol{\gamma}^{(\ell)},\, \boldsymbol{\beta}^{(\ell)}\bigr] &= \mathbf{W}_{\mathrm{f}}^{(\ell)}\,\mathbf{u} + \mathbf{b}_{\mathrm{f}}^{(\ell)}, \label{eq:rfd_film_params} \\
    \mathbf{h}_i^{(\ell+1)} &= \mathrm{ReLU}\!\Bigl(\boldsymbol{\gamma}^{(\ell)} \odot \mathcal{C}^{(\ell)}\bigl(\mathbf{h}_i^{(\ell)}\bigr) + \boldsymbol{\beta}^{(\ell)}\Bigr), \label{eq:rfd_film}
\end{align}
where $\mathcal{C}^{(\ell)}$ is the submanifold sparse convolution at layer $\ell$, and the FiLM scale and shift $\boldsymbol{\gamma}^{(\ell)}, \boldsymbol{\beta}^{(\ell)}$ are produced from a 16-dimensional global context $\mathbf{u}$ by a per-layer affine map with weight $\mathbf{W}_{\mathrm{f}}^{(\ell)}$ and bias $\mathbf{b}_{\mathrm{f}}^{(\ell)}$. The context $\mathbf{u}$ aggregates controller-velocity statistics (mean and norm), per-frame MfM output statistics (mean and standard deviation of $\log E$ and $\nu$), the normalized substep and frame phase via sinusoidal encoding, and the ground friction coefficient and height. A zero-initialized $1\times1\times1$ sparse convolution produces the residual, bounded by $\delta_{\max} = 0.05$\,m/s through the scaled $\tanh$ of Eq.~\eqref{eq:residual_update}, and RfD is applied every $H_{\mathrm{r}} = 10$ substeps.

\section{Material Augmentation Details}
\label{app:augmentation}

The material field of Eq.~\eqref{eq:material_field} is drawn from Perlin noise, a standard way to generate a smooth random field over space. Its value changes gradually from one point to the next, so nearby particles receive similar material rather than each being independently random. We add several higher-frequency octaves on top, so the field is heterogeneous at both coarse and fine scales.

For each episode we draw one such field for $\log E$ and an independent one for $\nu$. Each field is standardized to zero mean and unit variance, which separates its spatial pattern from its overall level and spread. The per-episode mean and standard deviation of Eq.~\eqref{eq:material_field} then set that level and spread, and we draw them independently from fixed uniform ranges. This produces elastic episodes with moderately heterogeneous stiffness and plastic episodes governed by a von Mises yield model.

\section{Training Details}
\label{app:training}

\paragraph{MfM pre-training.}
MfM is pre-trained for $2000$ epochs with batch size $4$, grouping episodes by source so each batch is homogeneous in particle count. The learning rate is held constant at $1\times10^{-4}$ throughout, with weight decay $0.01$. MfM is invoked every $K = 10$ frames, with backpropagation through time truncated every four windows. In Eq.~\eqref{eq:loss_refiner}, the confidence log-regularizer weight is $\lambda = 0.2$ and the plasticity term weight is $\beta = 0.1$. The plasticity term is also reweighted to balance the elastic-heavy dataset. For augmentation, each episode is randomly rotated about the gravity axis and mirror-flipped with probability $0.5$ along an orthogonal horizontal axis, applied to the canonical positions and controller points so the ground plane stays fixed. Tracked displacements and the controller inputs additionally receive $7$\,mm Gaussian jitter.

\paragraph{RfD training.}
RfD is optimized with truncated backpropagation through time. At each segment boundary we detach the final MPM state, so gradients stay within a single segment. The per-episode rotation and mirror flip above are also applied here, but to the full simulation state (positions, velocities $\mathbf{v}_p$, deformation gradients $\mathbf{F}_p$, and APIC affine matrices $\mathbf{C}_p$) because the RfD rollout evolves all of them.